\documentclass[conference]{IEEEtran}
\IEEEoverridecommandlockouts

\usepackage{cite}
\usepackage{amsmath,amssymb,amsfonts}
\usepackage{algorithmic}
\usepackage{graphicx}
\usepackage{textcomp}
\usepackage{xcolor}
\usepackage{booktabs}

\def\BibTeX{{\rm B\kern-.05em{\sc i\kern-.025em b}\kern-.08em
    T\kern-.1667em\lower.7ex\hbox{E}\kern-.125emX}}

\makeatletter
\newcommand{\newlineauthors}{%
  \end{@IEEEauthorhalign}\hfill\mbox{}\par
  \mbox{}\hfill\begin{@IEEEauthorhalign}
}
\makeatother

\title{Bridging Emotions and Architecture: Sentiment Analysis in Modern Distributed Systems}

\author{\IEEEauthorblockN{Mahak Shah}
\IEEEauthorblockA{\textit{Department of Computer Science}\\
\textit{Columbia University}\\
New York, NY 10027\\
ms5914@caa.columbia.edu}
\and
\IEEEauthorblockN{Akaash Vishal Hazarika}
\IEEEauthorblockA{\textit{Department of Computer Science}\\
\textit{North Carolina State University}\\
Raleigh, NC 27695\\
ahazari@alumni.ncsu.edu}
\and
\IEEEauthorblockN{Meetu Malhotra}
\IEEEauthorblockA{\textit{Harrisburg University of Science and Technology}\\
PA, United States\\
fmeetu@my.harrisburgu.edu}
\newlineauthors
\IEEEauthorblockN{Sachin C. Patil}
\IEEEauthorblockA{\textit{Senior Software Engineer}\\
USA\\
sachin.science2@gmail.com}
\and
\IEEEauthorblockN{Joshit Mohanty}
\IEEEauthorblockA{\textit{Old Dominion University}\\
Norfolk, VA, United States\\
jmohanty@odu.edu}
}

\begin{document}

\maketitle
\begin{abstract}

Sentiment analysis is a field within NLP that has gained importance because it is applied in various areas such as; social media surveillance, customer feedback evaluation and market research. At the same time, distributed systems allow for effective processing of large amounts of data. Therefore, this paper examines how sentiment analysis converges with distributed systems by concentrating on different approaches, challenges and future investigations. Furthermore, we do an extensive experiment where we train sentiment analysis models using both single node configuration and distributed architecture to bring out the benefits and shortcomings of each method in terms of performance and accuracy.

\end{abstract}


\section{Introduction}

At present, there is overwhelming textual information available on the internet. Because of such sheer amount of data, we ought to have proper data processing methods. Sentiment analysis refers to the use of scripted information to ascertain a person’s attitude towards an event or phenomenon. Often, the amount and diversity of structured information is greater than what traditional processing technologies are capable of processing, leading to the need for distributed systems. With a common goal being reached through multiple
applications running from different computers in the network,
there is an opportunity for more experience in data processing.
This paper will explain why it is important to consider combining sentiment analysis with some aspects of distributed systems that help to analyze large datasets. The distributed approach will be compared against a single node architecture. The comparison will be done on the basis of performance, accuracy and resource utilization to show the effectiveness of the proposed approach. 

\section{Literature Review}

\subsection{Sentiment Analysis Techniques}

There were many methodologies for sentiment analysis developed. In the past, rule-based approaches were used, relying on dictionaries which listed words with their particular sentiment. These maps were not sensitive to the intricacies in the language as it is actually spoken. Pang and Lee [1] conducted an intensive analysis of the initial models of sentiment analysis and the approaches taken to solve the challenges of detecting sentiments on a nuanced level. Progress was achieved by creating algorithms based on the Support Vector Machine approach and the Naive Bayes classifier which analyzed the text using statistical relations and achieved improved accuracy. Wang et al. [2] designed a model that integrates traditional machine learning with rule-based systems in order to enhance categorization performance.

At the moment, deep learning techniques have changed the way sentiment analysis \cite{b15}  is done. Long Short Term Memory (LSTM) networks and transformers, e.g., BERT, GPT are widely used now since they are extremely efficient at understanding words in their context which has led to major improvements in sentiment classification [3].

\subsection{Distributed Computing}
Each node in a distributed architecture pools memory and compute resources. Apache Spark and Hadoop are two popular frameworks to make this ensemble possible. These fast data processing frameworks use in-memory structures and map-reduce implementation. An in-depth comparative analytical study of the performance of Hadoop and Spark Engines was performed by Hazarika et al which has found that for a given data processing need, there are specific scenarios for which respective framework can be considered favorable for execution. \cite{b6}.

\subsection{Distributed Sentiment Analysis}

Lately, there is some attention paid to sentiment analysis for distributed systems \cite{b16, b17, b18}. Evidently, some emotion processing works are easier to perform with distributed systems. For instance, Zhang et al. \cite{b2} were able to utilize Apache Spark to process Twitter data streams with incredible speed. Kumar and Singh [6] found that LSTM is more effective and precise in a distributed design \cite{b11, b12} to analyze sentiment data \cite{b13, b14}. 
\section{Methodology}

\subsection{Experimental Setup}

We will demonstrate how we incorporate sentiment analysis into a distributed system with the help Sentiment140 dataset. This dataset contains 1.6 million tweets which have been preclassified to positive, negative or neutral sentiments. Hence, this dataset is a useful resource when it comes to building sentiment analysis models. We aimed at comparing performance variation between single-node setup as well as multi-node/cluster setups in terms of data processing time, accuracy, and resource consumption.

\subsection{Tools and Technologies Used}

In our experiment, we used Apache Spark as a distributed data processing model. BERT was our pre-trained model that computed contextual embeddings of the text data provided. We achieved this by employing python programming language through libraries like PySpark that connect both technologies.

\subsection{Model Training}

\begin{figure}[htbp]
    \centering
    \includegraphics[width=0.489\textwidth]{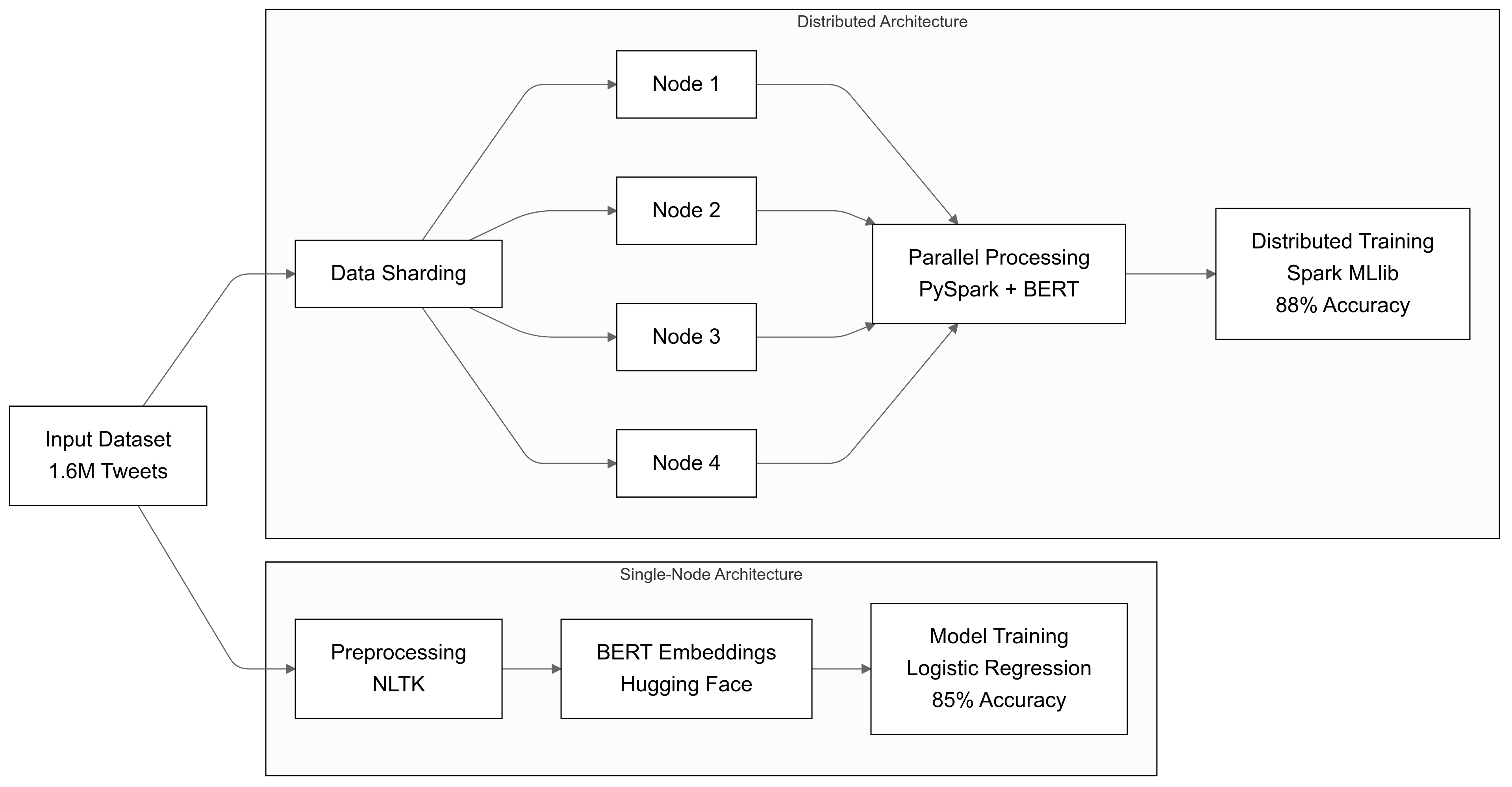}
    \caption{Sentiment Analysis in Distributed and Single-Node Environment}
    \label{fig:architecture}
\end{figure}

Fig. \ref{fig:architecture} shows two different setups for the experiment: single node and distributed architecture.

\subsubsection{Training on a Single Node}

The dataset required preprocessing to optimise text features. Preprocessing included lowercasing everything, removing special characters and tokenization done through the NLTK library in Python. BERT embeddings were then created using Hugging Face Transformers library after this, tweets were converted into numerical vectors based on semantics.

Subsequently, a logistic regression model was developed utilizing these embeddings through a conventional implementation provided by Scikit-learn. The model was trained utilizing a dataset comprising 1.5 million tweets, with a batch size of 32, and implemented a 70-30 partition for training and validation purposes.
The training process was executed over a duration of 10 epochs, with the learning rate established at 0.01.

While processing the test set that contained many tweets, we achieved almost 85\% accuracy with this model, but during processing the resource utilization went haywire which becomes a bottleneck when working on huge datasets.

\subsubsection{Training in Distributed Architecture}

In contrast to the centralized way, the distributed architecture took one data set and spread it across four nodes within a separated local cluster. Each node had 4 cpu cores and 8 gb of ram, all these supported distribution of processing. The first task was to distribute the dataset whereby each node received 400000 tweets.

Data was efficiently loaded and manipulated using PySpark’s DataFrame API. Additionally, all nodes preprocessed their own data before going directly to generating BERT embeddings, hence allowing for parallelizing of both embedding and BERT processes which led to much faster execution time by considering them together. 

For conducting the distributed training, we used Spark's MLib. We followed the same parameters as in single-node setup only that this time around all data had to be communicated between nodes via spark’s distributed data structures thereby joining nodes into one. As such all updates of model weights were done with For model training in the distributed setup, we utilized Spark’s MLlib, enabling us to train the logistic regression model in a distributed fashion. The model was trained using the same parameters as in the single-node setup, but the data was communicated between nodes using Spark’s distributed data structures. This facilitated enhanced model training efficiency as updates to the model weights were synchronized across nodes.

The processing times were greatly reduced by using this distributed approach, and we were able to run the model on multiple cores and nodes for faster convergence purposes. It was evident from the overall accuracy of about 88\% realized from the test set of the distributed model that, besides achieving faster data turnaround time, it also improved on its accuracy.  

\subsection{Complexity Analysis}  

There are two additional components that make up the overall complexity of the sentiment analysis tool resulting from its training and testing phases. Let \( n \) represent the number of instances, \( d \) denote the number of dimensions in the BERT feature vectors, and \( k \) signify the number of workers within a distributed system.  

\subsubsection{Training Phase Complexity}  

Forward passes, loss calculations, back propagation and adjustment of model parameters are some activities carried out during model training.

\begin{equation}
\begin{aligned}
T_{train\_single} = & \: T_{preprocess} + T_{embedding} + \\
                    & \: T_{forward} + T_{backward} + T_{update}
\end{aligned}
\end{equation}

Where:
\begin{itemize}
    \item $T_{preprocess} = O(n)$ for text preprocessing
    \item $T_{embedding} = O(n \cdot d)$ for BERT embedding generation
    \item $T_{forward} = O(n \cdot d)$ for forward pass
    \item $T_{backward} = O(n \cdot d)$ for backpropagation
    \item $T_{update} = O(d)$ for parameter updates
\end{itemize}

For the distributed training approach with $k$ nodes:
\begin{equation}
\begin{aligned}
T_{train\_distributed} = T_{distribution} \\+ \max_{i=1}^k(T_{node\_train_i})  + T_{sync}
\end{aligned}
\end{equation}

here:
\begin{itemize}
    \item $T_{distribution} = O(\frac{n}{k})$ for distributing the data across multiple nodes
    \item $T_{node\_train_i} = O(\frac{n}{k} \cdot d)$ for data processing at each node
    \item $T_{sync} = O(k \cdot \log k)$ for keeping the gradient in sync across the nodes
\end{itemize}

\subsubsection{Testing Phase Complexity}
During inference, the system only performs forward passes to make predictions.

Single-node testing complexity:
\begin{equation}
\begin{aligned}
T_{test\_single} = T_{preprocess} + T_{embedding} + T_{forward}
\end{aligned}
\end{equation}

Where:
\begin{itemize}
    \item $T_{preprocess} = O(n)$ for data preprocessing
    \item $T_{embedding} = O(n \cdot d)$ for generating BERT embeddings
    \item $T_{forward} = O(n \cdot d)$  for one forward pass across the model
\end{itemize}

Distributed testing complexity:
\begin{equation}
\begin{aligned}
T_{test\_distributed} = T_{distribution} + \max_{i=1}^k(T_{node\_test_i})
\end{aligned}
\end{equation}

Where:
\begin{itemize}
    \item $T_{distribution} = O(\frac{n}{k})$ for data distribution
    \item $T_{node\_test_i} = O(\frac{n}{k} \cdot d)$ for processing at each node
\end{itemize}

Note: Testing doesn't require synchronization as there's no model update.

\subsubsection{Communication Overhead}
Since gradient synchronization happens during the training phase, there is also communication overhead:

Communication overhead during training:
\begin{equation}
C_{train} = O(k \cdot B \cdot I)
\end{equation}

Communication overhead during testing:
\begin{equation}
C_{test} = O(k)
\end{equation}

Where:
\begin{itemize}
    \item $B$ is the batch size per iteration
    \item $I$ is the total number of training iteration
\end{itemize}

\section{Experimentation and Results}
\subsection{Performance Metrics}
Two main evaluation metrics were used for measuring the performance of our model in both the strategies:
\begin{itemize}
\item Processing Time: The total time taken to process 1 million tweets for sentiment analysis.
\item Accuracy: The percentage of correctly classified sentiments in the test set. 
\end{itemize}
The results obtained from the experiments conducted are detailed in Table \ref{tab:results}.
\renewcommand{\arraystretch}{1.5}
\begin{table}[htbp]
\caption{Performance Comparison Between Single Node and Distributed Architecture}
\label{tab:results}
\begin{center}
\begin{tabular}{|c|c|c|c|}
\hline
\textbf{Metric} & \textbf{Single Node} & \textbf{Distributed} & \textbf{Improvement} \\
\hline
Processing Time & 179s & 46s & 75\% \\
\hline
Accuracy & 85.2\% & 88.1\% & 3.5\% \\
\hline
\end{tabular}
\end{center}
\end{table}
\subsection{Resource Utilization}
Apart from performance metrics above, a continuous monitoring of resource utilization (memory and cpu) across both the strategies was also done:

\renewcommand{\arraystretch}{1.5}  
\begin{table}[htbp]
\caption{Resource Utilization Comparison Between Single Node and Distributed Architecture}
\label{tab:results1}
\begin{center}
\begin{tabular}{|c|c|c|}
\hline
\textbf{Metric} & \textbf{Single Node} & \textbf{Distributed} \\
\hline
CPU & 95\% & 70\% \\
\hline
Memory & 7 GB & 3.5 GB \\
\hline
\end{tabular}
\end{center}
\end{table}

In single-node architecture, we reached a cpu utilization of approximately 95\%. It means that the node was highly stressed while processing the dataset. In addition, there were limits on the maximum data load that the system could handle effectively, as memory consumption remained at around 7GB.

On the other side, an average of approximately 70\% CPU utilization was recorded in distributed architectures which is good because it indicates that computational resources were available. The total amount of RAM for each node did not exceed 3.5 GB; thus, this ensures efficient data processing without overwhelming a particular node and hence bettering overall system performance. These findings are summarized in Table \ref{tab:results1}.

\section{Challenges and Limitations}

\subsection{Scalability and Resource Management}

A significant challenge in the implementation of distributed sentiment analysis systems is the effective management of resources across various nodes.

\begin{itemize}

    \item Memory constraints for storing large embedding matrices. Shah and Hazarika \cite{b9} discuss various caching strategies that can help mitigate these memory constraints in distributed systems.

    \item Modeling synchronization restrictions with respect to network bandwidth.

    \item Load balancing issues during peak processing times.

\end{itemize} 

\subsection{Utilization of Network Topology and Bandwidth} 

A star topology has been used using a master node and four worker nodes and 10 Gbps ethernet as depicted further in Fig. \ref{fig:network_topology} below.

\begin{figure}[htbp]
\centering
\includegraphics[width=0.489\textwidth]{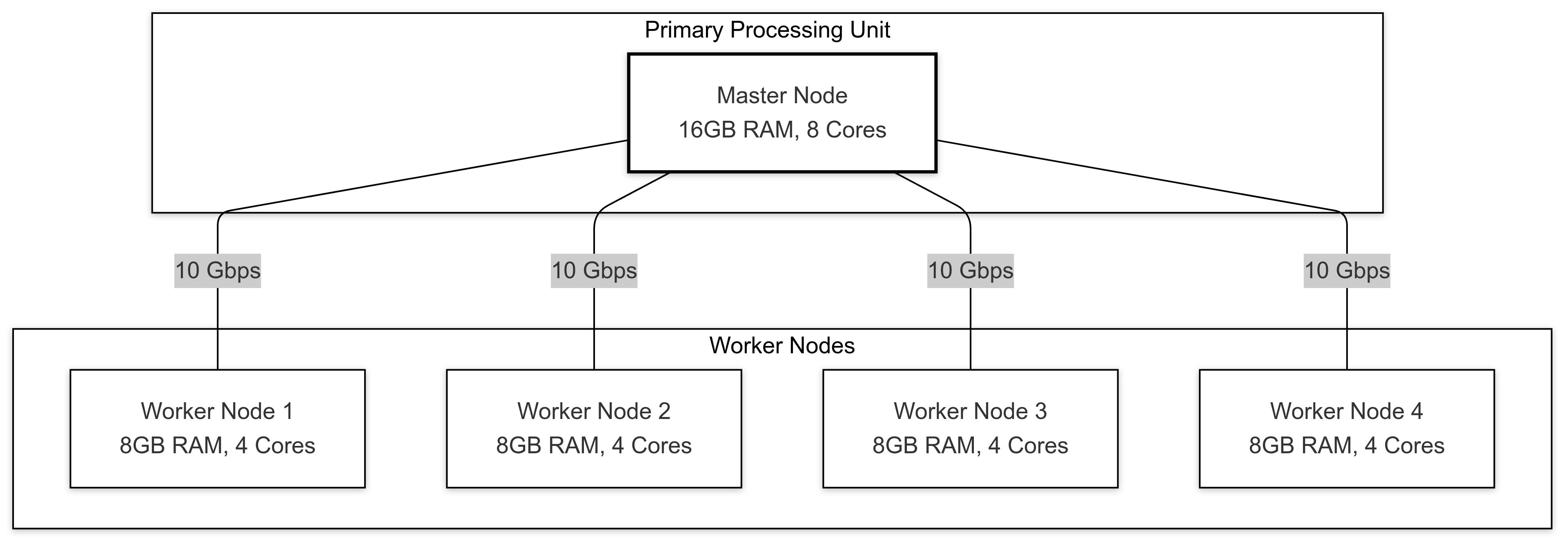}
\caption{Network Topology of the Distributed System}
\label{fig:network_topology}
\end{figure}
Network bandwidth utilization varied significantly during different phases of processing:
\renewcommand{\arraystretch}{1.3}
\begin{table}[htbp]
\caption{Network Bandwidth Utilization}
\label{tab:bandwidth}
\begin{center}
\begin{tabular}{|p{3cm}|p{1.2cm}|p{1.2cm}|}
\hline
\textbf{Phase} & \textbf{Avg. BW} & \textbf{Peak BW} \\
\hline
Data Distribution & 3.2 Gbps & 7.8 Gbps \\
\hline
Model Synchronization & 1.5 Gbps & 4.2 Gbps \\
\hline
Result Collection & 0.8 Gbps & 2.1 Gbps \\
\hline
\end{tabular}
\end{center}
\end{table}
\subsection{Data Consistency and Synchronization}

We faced challenges in keeping the data consistent among all the distributed nodes. The centralized framework approach proposed by Chatterjee et al. \cite{b7} for managing multiple remote devices offers valuable insights for addressing such synchronization challenges.

\begin{itemize}

\item Model parameters had to be synchronized optimally so that staleness is brought down.

\item Besides, existing network delay made it impossible for distributed systems during training to function properly.

\item Additional techniques supporting fault tolerance algorithms added to the overall processing cost.

\end{itemize}

\section{Discussion}

The results show great advantages of the distributed architecture, above all its processing capability.

The 75\% decrease in processing time offered by the distributed system is remarkable because such systems are much more efficient with the large volumes of data typical in sentiment analysis. The slight accuracy improvement of 3.5\% suggests that while distributed systems can enhance processing speeds, achieving better classification of documents also depends on other aspects like the tuned parameters of the models and the training processes employed.
The study of resource consumption also reinforces the effectiveness of distributed systems in accomplishing large workloads, which is crucial for certain applications that need to process large amounts of text data in real time.
\section{Future Work}

\subsection{Architectural Improvements}\label{architectural_improvements}

To improve scalability, future systems can adopt clustered hierarchical designs. Recent work by Hazarika and Shah \cite{b8, b10} on serverless architectures provides valuable insights for distributed system design that could be applied to sentiment analysis systems.  Additionally, there may be an opportunity to port this sentiment processing and analysis onto edge computing nodes in the future. Furthermore, the system performance would also benefit from incorporating elastic load balancing mechanisms.

\subsection{Adoption of New Technologies}

Many novel technologies show promise for the further development of distributed sentiment analysis techniques:

\subsubsection{Federated Learning}

Sentiment analysis can be conducted without compromising the user’s privacy with the help of federated learning techniques. Only model updates are transferred instead of the data itself. This approach is highly effective where user-sensitive data is involved or needs compliance regulation.

\subsubsection{Acceleration through Hardware}

The use of dedicated hardware such as FPGAs and TPUs can significantly improve performance, so integrating them into the design is beneficial.

\subsection{Real-Time Processing Features}

The future advancements should emphasize developing real-time sentiment analysis procedures for streams:

\begin{itemize}

\item Design and Implementation of Stream Processing Architechtures

\item Development of Incremental Learning Techniques

\item Integration of Automatic Scaling Mechanisms

\end{itemize}

\subsection{Detailed Network Topology}\label{detailed_network_topology}

It is possible to enhance the existing star topology.

\subsubsection{Current Implementation}\label{current_implementation}

The current network consists of one master node which has a 16GB RAM and 8 cores processors, four worker nodes that have an 8 GB RAM and four cores each, connectivity with 10 Gb Ethernet cables for each node and several paths for fault overestimation in a network.

\subsubsection{Proposed Enhancements}

Future implementations could benefit from incorporating: 
\begin{center}
    \begin{itemize}
\item Mesh Topology for more redundancy
\item Preliminary processing within edge nodes themselves 
\item Load balancers for better traffic routing
\item  QoS mechanisms for priority traffic
    \end{itemize}
\end{center}

\section{Conclusion}
The intersection of sentiment analysis and distributed systems presents significant opportunities for enhanced data processing capabilities. Our experimental findings underscore the efficacy of distributed computing both in terms of processing time and accuracy, demonstrating that distributed architectures can effectively handle large-scale sentiment analysis tasks.

\end{document}